# NEURAL ATTENTIVE MULTIVIEW MACHINES

*Oren Barkan[1*], Ori Katz[12*] and Noam Koenigstein[13]*

[1]Microsoft
[2]Technion
[3]Tel Aviv University

## ABSTRACT

An important problem in multiview representation learning is finding the optimal combination of views with respect to the specific task at hand. To this end, we introduce NAM: a Neural Attentive Multiview machine that learns multiview item representations and similarity by employing a novel attention mechanism. NAM harnesses multiple information sources and automatically quantifies their relevancy with respect to a supervised task. Finally, a very practical advantage of NAM is its robustness to the case of dataset with missing views. We demonstrate the effectiveness of NAM for the task of movies and app recommendations. Our evaluations indicate that NAM outperforms single view models as well as alternative multiview methods on item recommendations tasks, including cold-start scenarios.

## 1. INTRODUCTION AND RELATED WORK

Multiview representation learning [22] is a rapidly growing research field concerned with the task of learning entity (item) representations [32] and similarities from several different information sources (views). The learned representations extract useful information from the different views and are later used for inferring *multiview* item similarities.

The explosion of readily available data in companies and organizations generates many opportunities to leverage various data sources for learning multiview item representations [1], [3], [22], [28]. For example, a movie (item) can be represented by its audio-visual representation (e.g. video, soundtrack), textual description (plot summary), tags descriptions (actors, genres etc.) or Collaborative Filtering (CF) information (e.g. other movies that were co-consumed with the specific movie). Each data source enables a different *view* of the same item.

Different views however, vary in resolution and usefulness. For example, the genre view of a movie is usually less informative than its CF view, since inferring movies relations at the genre level is very coarse. A good multiview model learns a multiview similarity function that integrates information from each view based on its usefulness w the other views and the specific task at hand.

In real world datasets, it is often the case that different views cover different subsets of items. For example, when a new movie is released, CF information is not available and item relations can only be inferred by Content Based (CB) information (e.g. genres, actors, plot, etc.). In Recommender Systems research, this is known as the items *cold-start* problem [23]. A good multiview model can graciously fallback to a similarity function based on available views (e.g. CB views) when CF information is missing. Hence, the multiview approach provides substantial merit when working with missing data.

NAM is an attention based model that learns item representations and similarities based on multiple views. First, view-dependent item representations are learned in a latent vector space with a view dependent similarity function. Then, the final multiview similarity score is obtained via a novel neural attention mechanism that learns to quantify the relative contribution from each view w.r.t. the other views and the task at hand. In this paper, we demonstrate NAM using datasets of movies and apps containing both CB and CF data. Evaluations indicate that NAM produce multiview item similarities that are superior to other singleview and multiview models, and gracefully handles cold start scenarios.

Neural attention mechanisms are emerging techniques that affects a variety of different fields [2], [11], [19], [21], [31], [32], [33]. For instance, in [24] the authors suggest using attention mechanism for improving the nearest neighbors in each view, however, the integration between different views is set by hyperparameters. In [8] the authors utilize attention mechanism with multiview data, but the proposed model is not originally designed for handling cold-start scenarios. Different from [24] and [8], NAM produces multiview item similarities aside with attention mechanisms. Another key feature that differentiates NAM from the aforementioned works is its ability to produce item similarities in missing view scenarios.

The problem of multiview representation learning is an active research field [1], [3], [13], [22], [29]. In the context of Recommender Systems, Collaborative Deep Learning (CDL) [28] is a very popular hybrid recommendations model. CDL employs a denoising autoencoder along with WMF. Recently, CB2CF [6] was proposed for addressing the cold-start problem in recommender systems. CB2CF, can be seen as an out-of-sample extension [34] technique that learns a mapping from a CB view of items to the CF view via a deep multiview model. Instead, NAM utilizes any available views to achieve its ultimate task (e.g., learning item recommendations). In Section 3, we show that NAM is comparable to CB2CF in the cold start scenarios, while outperforming other existing methods when CF data is available.

Our contribution is in introducing a novel model for attentive multiview item similarity. The model naturally handles missing views and provides competitive results on recommendation tasks including cold start scenarios.

## 2. NEURAL ATTENTIVE MULTIVIEW MACHINES

NAM is a general attentive multiview model that can be employed on a plethora of supervised tasks. In Sections 2.1 and 2.2, we generally explain the NAM model. Next, in Section 2.3 we provide further details on a version of NAM specifically designed for multiview item recommendations. This particular version of NAM is evaluated in Section 3.

---
[*]Equal contribution

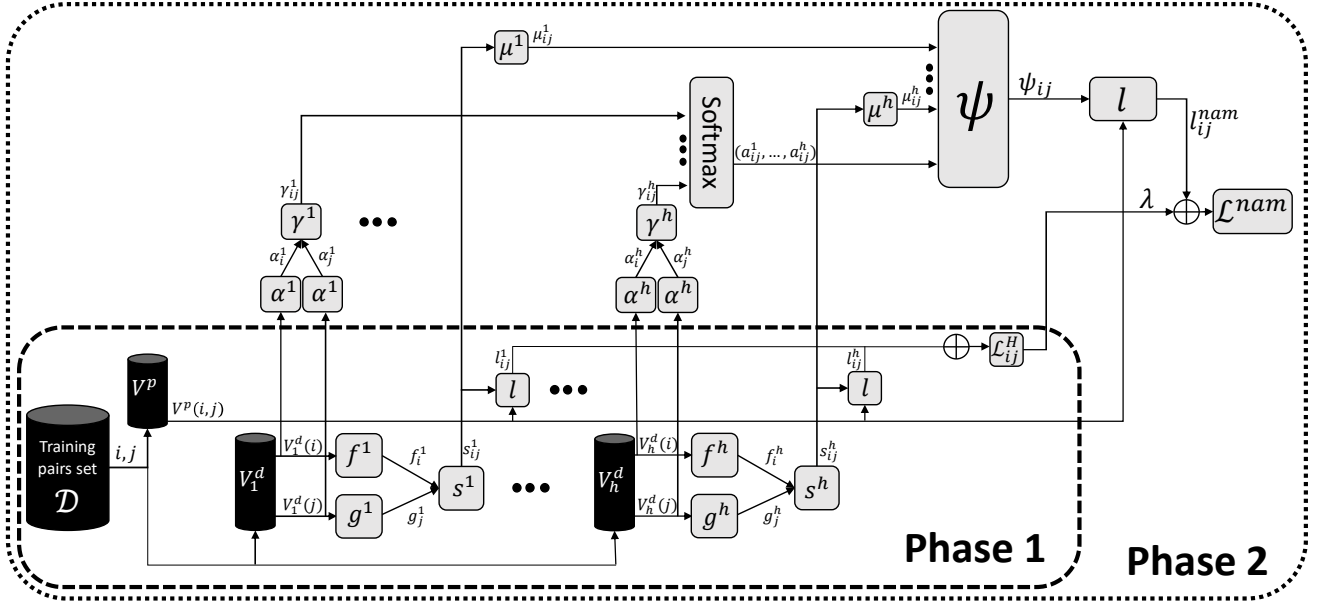

Figure 1. A schematic illustration of NAM

Let $I = \{i\}_{i=1}^K$ be a set of items. A *direct* view $V^d: I \to \mathbb{R}^k$ of $I$ is a function that maps an item to a vector. Let $V_H = \{V_h^d\}_{h \in H}$ be a set of explaining views, where $h \in H$ is an index that represents the view type. The explaining views $V_H$ provide different complementary information about the items. For example, if $h$ represents textual (visual) information, then $V_h^d(i)$ is a vector that encodes the textual (visual) description of the item $i$.

A *pairwise* view of $I$ is a function $V^p: I \times I \to \mathbb{R}$ that receives a pair of items $(i,j)$ and outputs a scalar. $V^p$ is used as a *supervisor* that determines the true affinity between pairs of items. For example, in the task of collaborative filtering dataset, the supervision is determined by item co-occurrences. Hence, $V^p(i,j) = 1$ if the items $i$ and $j$ were co-consumed by a user, otherwise $V^p(i,j) = 0$.

Let $\mathcal{D}$ be a dataset of item pairs $(i,j) \in \mathcal{D}$. Then, for each pair $(i,j)$, NAM is supervised by the supervisor view that produces labels $V^p(i,j)$. For example, in the case of item co-occurances, $\mathcal{D}$ contains positive item-pairs $(i,j)$ that were co-consumed by the same user ($V^p(i,j) = 1$) and negative item-pairs $(i,n)$ that were never co-consumed together by the same user ($V^p(i,j) = 0$).

NAM is employs a two-phase process: In the first phase, the model learns view-dependent item representations and corresponding similarity functions by using $\mathcal{D}$ and the supervisor view $V^p$. Then, in the second phase, $\mathcal{D}$, $V^p$ and the view-dependent similarity functions are used for producing a final multiview attentive scoring mechanism. Next, we describe the general NAM model and its two-phase training procedure. The NAM architecture is depicted in Fig. 1 and detailed below.

### 2.1. Phase 1 – View Dependent Training

In the first phase, NAM learns view-dependent representations and similarity functions. To this end, for each view, $V_h^d$ we define view-dependent *context* and *target* embedding functions $f^h, g^h: \mathbb{R}^{z_h} \to \mathbb{R}^{z_t}$, where $z_h$ stands for the dimension of the direct view $V_h^d$. We denote by $f_i^h \triangleq f^h(V_h^d(i))$ and $g_i^h \triangleq g^h(V_h^d(i))$ the mapping of the different (direct) item views $V_h^d(i)$ into two $z_t$-dimensional vector spaces. Namely, $f^h, g^h$ map item $i$'s direct views $V_h^d(i)$ (e.g. audio, image, text) into $f^h$ and $g^h$, the context and target latent representations respectively.

Let $s^h: \mathbb{R}^{z_t} \times \mathbb{R}^{z_t} \to \mathbb{R}$ be a view-dependent scoring function that computes the similarity for item-pair $(i,j)$ using their corresponding context and target representations $f_i^h$ and $g_j^h$. The similarity score of an item-pair $(i,j)$ induced by $s^h$ is denoted by $s_{ij}^h \triangleq s^h(f_i^h, g_j^h)$.

Let $l: \mathbb{R} \times \mathbb{R} \to \mathbb{R}$ be a loss function that computes the loss for (score, label) pairs, where the score and label are obtained by the scoring function $s^h$ and the supervisor view $V^p$. In addition, denote $\mathcal{L}_{ij}^H \triangleq \frac{1}{|H|} \sum_{h \in H} l_{ij}^h$ with $l_{ij}^h \triangleq l\left(s_{ij}^h, V^p(i,j)\right)$. Then, our objective is to minimize,

$$\mathcal{L}^H = \sum_{(i,j) \in \mathcal{D}} \mathcal{L}_{ij}^H. \quad (1)$$

Note that when minimizing $\mathcal{L}_{ij}^H$, each term $l_{ij}^h$ is minimized independently and hence $f^h, g^h$ and $s^h$ are also learned independently for each view in separate.

Once $f^h, g^h$ and $s^h$ are learned for each view, we proceed to the second training phase that performs attentive multiview training.

### 2.2. Phase 2 – Attentive Multiview Training

In this phase, NAM learns an attentive multiview scoring function that combines the view-dependent similarity scores. This scoring function utilizes an attention mechanism that is designed to quantify the contribution of each view $h$ to the similarity between items $i$ and $j$, and handles missing views.

Let $\alpha^h: \mathbb{R}^{z_h} \to \mathbb{R}^{z_a}$ be a view-dependent attention function, where $h \in H$ and denote $\alpha_i^h \triangleq \alpha^h\left(V_h^d(i)\right)$. $\alpha_i^h$ maps the view representation of item $i$ to a $z_a$- dimensional attention vector. For each view $h$, the direct views $V_h^d(i)$ and $V_h^d(j)$ of items $i$ and $j$ are mapped to $\alpha_i^h$ and $\alpha_j^h$, respectively.

Let $\gamma^h: \mathbb{R}^{z_a} \times \mathbb{R}^{z_a} \to \mathbb{R}$ be a view dependent pairwise attentive scoring function and denote $\gamma_{ij}^h \triangleq \gamma^h(\alpha_i^h, \alpha_j^h)$. $\gamma_{ij}^h$ quantifies the amount of useful (pairwise) information provided by the view $h$. Then, the multiview attentive similarity score is given by $\psi_{ij} = \sum_{h \in H} a_{ij}^h \mu_{ij}^h$, where $\mu_{ij}^h = w^h s_{ij}^h + b^h$ is a view-dependent affine transformation designed to normalize the view-dependent similarity score, and $a_{ij}^h = \exp(\gamma_{ij}^h) / \sum_{h \in H} \exp(\gamma_{ij}^h)$ are the attention coefficients. $\psi_{ij}$ is the final output of the model. Therefore, during inference, NAM scores a pair of items $(i, j)$ using $\psi_{ij}$. Note that the output $\psi_{ij}$ is a combination determined by the view dependent attention functions $a_{ij}^h$ that considers the relative information coming from a specific view $h$ with respect to all other views.

Finally, the NAM loss at the second phase of training is defined as

$$\mathcal{L}^{nam} = \sum_{(i,j) \in \mathcal{D}} l_{ij}^{nam} + \lambda \mathcal{L}_{ij}^H \qquad (2)$$

with $l_{ij}^{nam} = l(\psi_{ij}, V^p(i, j))$, and $\mathcal{L}_{ij}^H$ as defined earlier. The final NAM loss $\mathcal{L}^{nam}$ balances between learning the optimal attentive combination of views by minimization of the loss on the output terms $\psi_{ij}$ while still allowing small amount of finetuning $f^h, g^h$ and $s^h$ via the $\lambda \mathcal{L}_{ij}^H$ loss term.

### 2.3. Attentive Multiview Item Recommendations

The NAM model is a general multi-view framework that can be employed for a variety of tasks. Next, we present a specific version of NAM designed for multiview item recommendations. As explained above, in this particular case, the data consists of item pairs $(i, j)$ that were co-consumed by the same user. Note that $i$ and $j$ might be co-consumed more than once by different users and hence $(i, j)$ may appear in the data multiple times. For each positive example $(i, j)$ we randomly draw $N$ negative examples $\{(i, n_k)\}_{k=1}^N$ such that $(i, n_k)$ were never co-consumed together by the same user. These positive and negative pairs define the dataset $\mathcal{D}$. The pairwise supervisor view $V^p$ maps each positive and negative pairs to 1 and 0, respectively.

The explaining views we use are $V_H = \{V_{CF}^d\} \cup \{V_{C_m}^d\}_{m \in M}$. $V_{CF}^d$ is the latent collaborative filtering view of the items obtained by item2vec [7], and $\{V_{C_m}^d\}_{m \in M}$ are content based views derived from different types of metadata that exist for the items. Specifically, for movies we used the following views: genres, actors and tags (each represented a binary vector), year of release (a positive integer), and latent vector that encodes the movie textual description (computed by a pretrained BERT model [10] that operates on the raw text). For apps, we used tags (binary vector), category (one-hot vector) and the app textual description (encoded using BERT in the same manner as done for the movies).

For all $m \in M$, $f^{C_m}$ and $g^{C_m}$ are implemented as fully connected (FC) neural networks with a single ReLU activated hidden layer of size $z_t$. $f^{CF}, g^{CF}$ and $\alpha^h$ are implemented as FC linear networks. The scoring functions $s^h$ and pairwise attentive functions $\gamma^h$ are implemented as the cosine similarity that receives $f^h, g^h$ and $\alpha^h, \alpha^h$ as input, respectively, and outputs a scalar.

The loss function $l$ is set to the negative log softmax with negative sampling (SNS). This loss function was successfully applied for item recommendations in [7]. By setting the SNS loss in Eqs. (1) and (2) we get

$$l_{ij}^h = -s_{ij}^h + \log \sum_{k=1}^N \exp(s_{in_k}^h)$$

and

$$l_{ij}^{nam} = -\psi_{ij} + \log \sum_{k=1}^N \exp(\psi_{in_k}),$$

respectively. Both $\mathcal{L}^H$ (phase 1) and $\mathcal{L}^{nam}$ (phase 2) are minimized via stochastic gradient descent.

As explained above, NAM uses all the available views to produce the ultimate pairwise scores given the specific task at hand. In case of cold items, NAM naturally scores a pair of items $(i, j)$ via $\psi_{ij}$, but using the available views only. Thanks to the attention mechanism, handling cold items is done by setting the pairwise attention coefficient $\gamma_{ij}^{CF}$ to $-\infty$ if $i$ or $j$ are cold, respectively. For example, assume $i$ is a warm item and $j$ is a cold item and lacks CF representation. Then, $\gamma_{ij}^{CF} = -\infty \to a_{ij}^{CF} = 0$.

## 3. EXPERIMENTAL SETUP AND RESULTS

Our evaluation includes two datasets: First is the Movies dataset is based on the publicly available MovieLens dataset [14] consisting of both CF data and CB. It consists of 22,884,377 ratings collected from 247,753 users that watched some 34,208 movies. The movies are rated using a 5-star scale and for each user we consider all the movies with ratings above 3.5 to produce a dataset of co-occurring movies. This results in 11,108 unique movies (items) consumed by 173,266 users (sets), which are used to form the CF view with item2vec [7]. For each movie, we further collected metadata from TMDB and IMDB to form the genres, actors, year, tags and textual description views as explained in Section 2.3.

The second dataset is a propriety dataset containing CF and CB data of apps from the Microsoft Windows Store. The CF view is obtained by the application of item2vec to a sample of 33K unique items (apps) and 5M user activity sessions. Each session is a list of items that were co-clicked during the same session. The CB views (tags, category, textual description) are obtained as explained in Section 2.3.

The evaluation covers four models: two versions of NAM and two other baselines as follows:

**I2V**: This is the item2vec model from [7]. This model is used as a CF baseline model. I2V cannot produce recommendations for cold items as it uses CF relations only.

**CB2CF**: This model was recently proposed in [6] for mitigating the cold start problem. It employs deep regression from items CB representations to their corresponding CF representations. Then, it is used to predict the CF vectors for cold items. CB2CF optimized for handling cold start scenarios and serves as a challenging baseline for evaluating NAM in cold start scenarios.

**NAM**: The proposed model (Section 2).

**NAM-CB**: This is a modified version of NAM that utilizes the content based views only $\{V_{C_m}^d\}_{m \in M}$. In this version, we omit the CF view $V_{CF}^d$ by setting $\gamma_{ij}^{CF} = -\infty$ for all $(i, j)$.

All hyperparameters were determined using a separated validation set. The target dimension for I2V, CB2CF and NAM variants ($z_t$) was set to 100. For NAM we set $\lambda = 0.1$. All models were trained for 60 epochs using ADAM [17] optimizer with minibach size of 32. Negative sampling ratio was set 4 for all models that use negative samples.

We report mean values obtained from 10 fold cross validation procedure. The following evaluation measures are considered:

Hit Ratio **HR@K** [23]: For a positive pair $(i, j)$ HR@K outputs 1 if target item $j$ is ranked in the top K recommendations by the model w.r.t. the query item $i$ and 0 otherwise.

Mean Reciprocal Rank (**MRR@K**) [23]: This measure assigns a monotonic decreasing score for the rank of the target item if it is among the top K recommendations and 0 otherwise. Different from HR@K, the MRR@20 measure does consider the order of the recommendation list.

Figure 2 depicts HR@K and MRR@K results for the Movies and Apps datasets, respectively. These results showcase the ability of

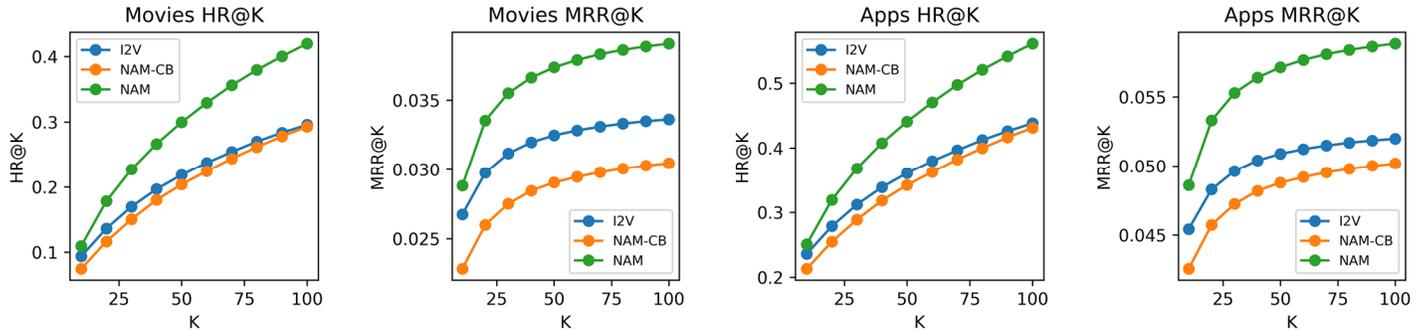

Figure 2. HR@K and MRR@K for the Movies and Apps datasets.

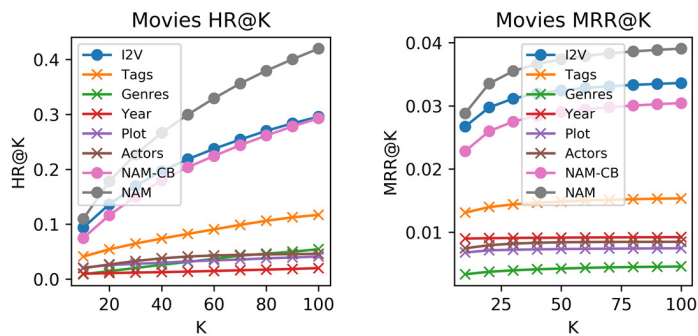

Figure 3. Ablation study. HR@K and MRR@K results for Movies.

NAM to generate superior recommendations when compared to the considered baselines.

Figure 3 depicts the HR@K and MRR@K curves for an ablation study over the views, on the Movies dataset (due to space limitation, we do not present an ablation study for the Apps datasest). We can see that NAM is the champion and NAM-CB is competitive with I2V (CF view), even though it utilizes CB views only. Among each of the individual CB views, the tags view produces the best result. This can be explained by the rich and informative set of tags that is associated with the movies.

Although NAM is not trained specifically for solving cold start scenarios, handling cold items is an inherent feature of NAM. We compare NAM to CB2CF which is specifically designed for addressing the cold start scenario and show competitive results. The following experiment focuses on the case in which some of the items are cold with no CF data at all. Consider a catalog of items denoted as the warm catalog. In this catalog, each item has at least one acquisition. Consider another new set of cold items that have no CF data at all, denoted as the cold catalog. We evaluate HR@20 and MRR@20 for each one of the following two scenarios:

**Warm:** Item-to-item (I2I) recommendations within the warm catalog (both items belong to the warm catalog). This test case includes item-pairs $(i, j)$ such that both $i$ and $j$ are warm.

**Cold:** I2I recommendations in which one of the items belongs to the cold catalog, and the other may belong to the cold or the warm catalog. The cold test set includes item-pairs $(i, j)$ of three different cases: 1) $i$ and $j$ are both cold. 2) $i$ is warm and $j$ is cold. 3) $i$ is cold and $j$ is warm.

In the *warm* scenario, we evaluate the ability of NAM to identify recommendations within the existing warm catalog which does not exhibit any degradation due to the introduction of new cold items. The second *cold* scenario, evaluates the ability of NAM to provide recommendations when either the query or the candidate items are cold. For each dataset, we perform 10 folds cross validation according to the following procedure: We simulate the cold test catalog by randomly picking 10% of the items and discarding their acquisitions from the train users (or session in the Apps dataset) history. Then, each model was trained on the remaining 90% of the items which now forms the "warm" train catalog. We then merged the two catalogs and calculated the resulting evaluation measures, separately for each scenario, on the test users (or sessions in the Apps dataset).

The HR@20 and MRR@20 results for the Movies and Apps datasets are presented in Tables 1 and 2, respectively. Recall that CB2CF addresses only the cold start scenario. Hence, it is evaluated on the cold catalog only. NAM is superior at handling cold start scenarios without causing any degradation to the warm catalog. NAM remains superior within all of the tested scenarios and therefore can serve as a hybrid system while supporting cold start scenarios as well. Moreover, NAM performs slightly better than CB2CF which is designed especially for the cold start scenario.

Table 1: "Warm" and "Cold" HR@20 values per model the Movies and Apps datasets.

|  | Movies | | | Apps | | |
| --- | --- | --- | --- | --- | --- | --- |
|  | NAM-CB | CB2CF | NAM | NAM-CB | CB2CF | NAM |
| Warm | 0.12 | - | **0.177** | 0.276 | - | **0.321** |
| Cold | 0.088 | 0.097 | **0.11** | 0.171 | 0.174 | **0.181** |

Table 2: "Warm" and "Cold" MRR@20 values per model on the Movies and Apps datasets.

|  | Movies | | | Apps | | |
| --- | --- | --- | --- | --- | --- | --- |
|  | NAM-CB | CB2CF | NAM | NAM-CB | CB2CF | NAM |
| Warm | 0.027 | - | **0.0339** | 0.0478 | - | **0.0537** |
| Cold | 0.0172 | 0.0176 | **0.018** | 0.0265 | 0.029 | **0.031** |

We remark two interesting observations regarding NAM-CB. We notice that NAM outperforms NAM-CB on the cold scenarios. This behavior is originated in two reasons: The first reason is due to the fact that the cold scenario includes three different cases as explained above. In case 2 ($i$ is warm and $j$ is cold), the context item $i$ is warm and hence has a CF view. Therefore, NAM utilizes the available CF view when computing the scores between $i$ and the rest of the warm items in the catalog. In contrast, NAM-CB utilizes the CB views only. The second reason is due to a gap between validation and training errors in NAM-CB. This gap can be noticed by comparing the performances of NAM-CB between the warm and cold regimes, demonstrating the inherent trade-off between providing warm to warm recommendations and supporting cold start scenarios.

## 4. CONCLUSION AND FUTURE WORK

We presented NAM – a neural attentive multiview machine that facilitates multiview item similarity with a novel attention mechanism. NAM learns to attend the relevant views that most contribute to the task at hand. Moreover, NAM naturally produces multiview item similarities, where some of the views are missing. We demonstrated the application of NAM for movies and apps recommendations and showcased its ability to learn multiview item similarities from multiple CB and CF views. In empirical evaluation, NAM outperformed other existing baselines across various measures, both in warm regimes and cold start scenarios.

In the future, we plan to employ NAM for other tasks, where the item-pairs $(i, j)$ and hence $\mathcal{D}$ are derived differently. For example, NAM can be used for identity recognition and verification tasks [4], [5] by compiling a dataset of pairs $(i, j)$, where $i$ and $j$ are two examples of the same identity (person), and $(i, n)$ are two examples of different identities. In this case, the different views can represent different images, voice samples or any additional auxiliary information available on the identities. Then, NAM utilizes the multiple views for performing multifactor biometrics.